\def\cvprPaperID{5955}
\def\confName{CVPR}
\def\confYear{2023}

\def\paperTitle{Uncertainty-aware Vision-based Metric Cross-view Geolocalization}

\def\authorBlock{
	\newcommand{\spacing}{\hspace{0.3mm}}
	\hspace{-3mm}
    Florian Fervers$^1$ \spacing
    Sebastian Bullinger$^1$ \spacing
    Christoph Bodensteiner$^1$ \spacing
    Michael Arens$^1$ \spacing
    Rainer Stiefelhagen$^2$ \\
    $^1$Fraunhofer IOSB \quad $^2$Karlsruhe Institute of Technology \\
    {\tt\small $^1$\{firstname.lastname\}@iosb.fraunhofer.de \quad $^2$rainer.stiefelhagen@kit.edu}
}

\newif\ifreview 
\newif\ifarxiv 
\newif\ifcamera \newcommand{\cameraready}{\cameratrue}
\newif\ifrebuttal 

\cameraready %

\pdfoutput=1
\documentclass[10pt,twocolumn,letterpaper]{article}
\ifreview \usepackage[review]{cvpr} \fi
\ifarxiv \usepackage[pagenumbers]{cvpr} \fi
\ifrebuttal \usepackage[rebuttal]{cvpr} \fi
\ifcamera \usepackage{cvpr} \fi

\usepackage{graphicx}
\usepackage{amsmath}
\usepackage{amssymb}
\usepackage{booktabs}

\usepackage{times}
\usepackage{microtype}
\usepackage{epsfig}
\usepackage[table,xcdraw]{xcolor}
\usepackage{caption}
\usepackage{float}
\usepackage{placeins}
\usepackage{color, colortbl}
\usepackage{stfloats}
\usepackage{enumitem}
\usepackage{tabularx}
\usepackage{xstring}
\usepackage{multirow}
\usepackage{xspace}
\usepackage{url}
\usepackage{subcaption}
\usepackage{xcolor}
\usepackage[hang,flushmargin]{footmisc}

\ifcamera \usepackage[accsupp]{axessibility} \fi

\ifarxiv  \fi

\newcommand{\R}[1]{{%
    \textbf{%
        \ifstrequal{#1}{1}{\textcolor{red}{R#1}}{%
        \ifstrequal{#1}{2}{\textcolor{blue}{R#1}}{%
        \ifstrequal{#1}{3}{\textcolor{magenta}{R#1}}{%
        \ifstrequal{#1}{4}{\textcolor{teal}{R#1}}{%
                           \textcolor{cyan}{R#1}%
        }}}}%
    }%
}}

\usepackage{xr-hyper}

\makeatletter
\newcommand*{\addFileDependency}[1]{
  \typeout{(#1)}
  \@addtofilelist{#1}
  \IfFileExists{#1}{}{\typeout{No file #1.}}
}

\makeatother

\usepackage[pagebackref,breaklinks,colorlinks]{hyperref}
\usepackage[capitalize]{cleveref}
\crefname{section}{Sec.}{Secs.}
\crefname{table}{Table}{Tables}
\crefname{figure}{Fig.}{Figs.}

\usepackage{amssymb}
\usepackage{pifont}
\usepackage{float}
\usepackage{capt-of,etoolbox}

\usepackage{xcolor}
\usepackage{arydshln}

\newcommand{\cmark}{\ding{51}}%
\newcommand{\xmark}{\ding{55}}%

\newcommand*{\tran}{^{\mkern-1.5mu\mathsf{T}}}

\begin{document}

\makeatletter
\let\@oldmaketitle\@maketitle
\renewcommand{\@maketitle}{
\@oldmaketitle
\myfigure
\bigskip
}
\makeatother

\newcommand\myfigure{
	\centering
	\vspace{-1mm}
	\includegraphics[width=0.97\linewidth]{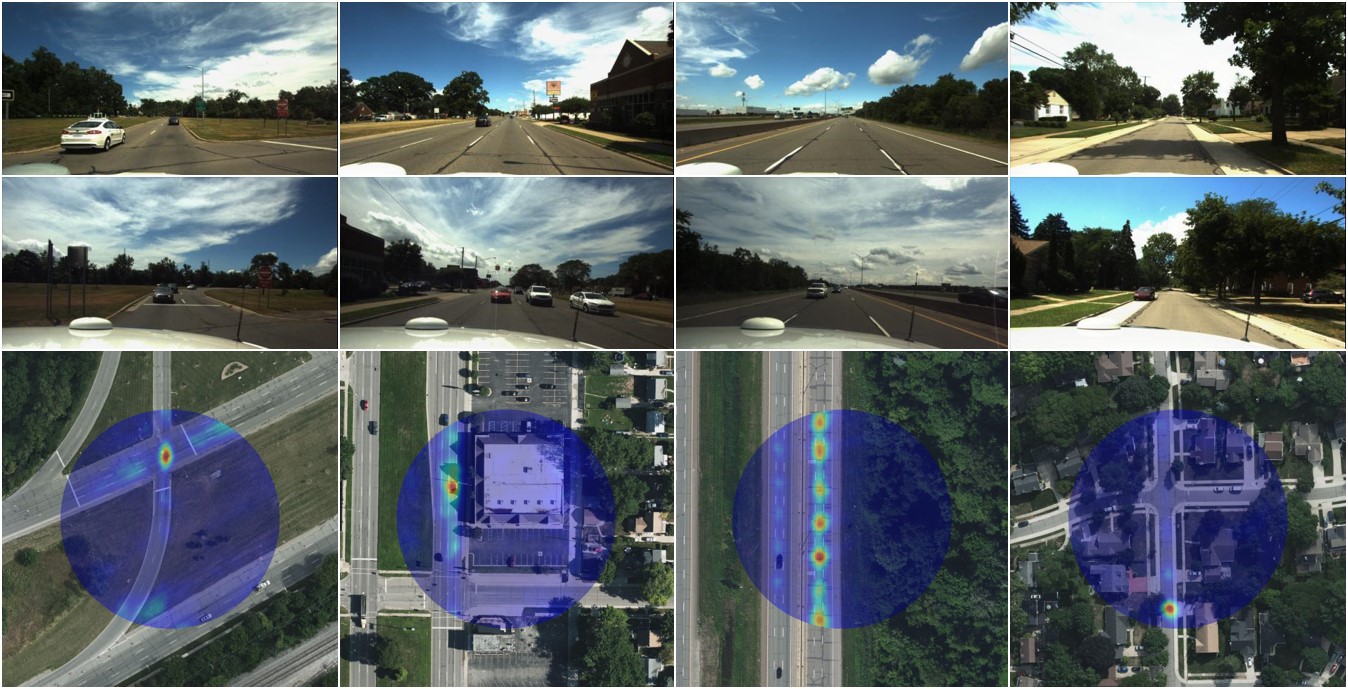}
	\vspace{-2mm}
	\captionof{figure}{Probability distributions for the vehicle position predicted by our model which matches the vehicle's surround camera images with an aerial image. The first and second rows show the front and back cameras in the Ford AV dataset \cite{agarwal2020ford}. The last row shows the aerial image with the search region in the center and driving direction pointing upwards. Blue and red color refer to low and high probability predicted by our model. Map data: Bing Maps © 2022 TomTom, © Vexcel Imaging \cite{bingmaps}.}
	\label{fig:intro}
}

\title{\paperTitle}
\author{\authorBlock}
\maketitle

\begin{abstract}
\setlength{\parskip}{0cm}

This paper proposes a novel method for vision-based metric cross-view geolocalization (CVGL) that matches the camera images captured from a ground-based vehicle with an aerial image to determine the vehicle's \mbox{geo-pose}. Since aerial images are globally available at low cost, they represent a potential compromise between two established paradigms of autonomous driving, \ie using expensive \mbox{high-definition} prior maps or relying entirely on the sensor data captured at runtime.

We present an end-to-end differentiable model that uses the ground and aerial images to predict a probability distribution over possible vehicle poses. We combine multiple vehicle datasets with aerial images from orthophoto providers on which we demonstrate the feasibility of our method. Since the ground truth poses are often inaccurate \wrt the aerial images, we implement a pseudo-label approach to produce more accurate ground truth poses and make them publicly available.

While previous works require training data from the target region to achieve reasonable localization accuracy (\ie \emph{same-area} evaluation), our approach overcomes this limitation and outperforms previous results even in the strictly more challenging \mbox{\emph{cross-area}} case. We improve the previous state-of-the-art by a large margin even without ground or aerial data from the test region, which highlights the model's potential for global-scale application. We further integrate the \mbox{uncertainty-aware} predictions in a tracking framework to determine the vehicle's trajectory over time resulting in a mean position error on KITTI-360 of 0.78m.

\end{abstract}

\if{false}
\begin{figure}[t]
	\centering
	\includegraphics[width=\linewidth]{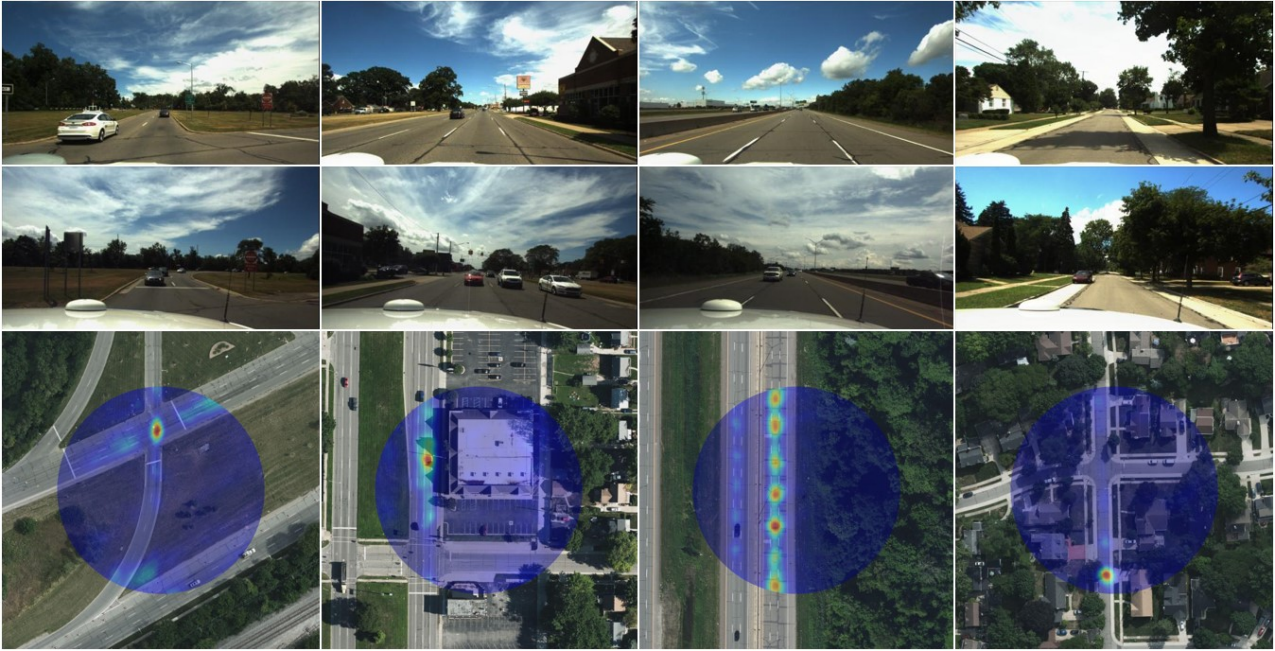}
	
	\caption{Probability distributions for the vehicle position predicted by our model which matches the vehicle's surround camera images with an aerial image. The first and second rows show the front and back cameras in the Ford AV dataset \cite{agarwal2020ford}. The last row shows the aerial image with the search region in the center and driving direction pointing upwards. Blue and red color refer to low and high probability predicted by our model. Map data: Bing Maps © 2022 TomTom, © Vexcel Imaging \cite{bingmaps}.}
	\label{fig:intro}
\end{figure}
\fi

\section{Introduction}
\label{sec:intro}

Systems for autonomous driving require both a model of the vehicle's environment as well as the location of the vehicle relative to the model. These systems either construct the full model during runtime (\ie entirely online), or create some parts prior to runtime (\ie partly offline). The latter methods typically construct \mbox{high-definition} maps of a region in advance (\eg using lidar sensors) and localize the vehicle at runtime relative to that map \cite{rozenberszki2020lol}. While prior maps facilitate a high localization accuracy of the system, they are also expensive to construct and maintain. Online methods on the other hand create a model of the local environment using only the live sensor readings \eg from lidar \cite{ye2022lidarmultinet}, camera \cite{harley2022simple} or both \cite{liu2022bevfusion}. This avoids the need for expensive prior maps, but represents a more difficult task as the system has to predict both the spatial structure of the environment as well as its relative location within it.

Aerial images offer the potential to leverage the advantages of both approaches: They can be used as a prior map for localization, while also being affordable, globally available and up-to-date due to an established infrastructure of satellite and aerial orthophoto providers \cite{googlemaps,bingmaps}. We consider the problem of matching the sensor measurements of the vehicle against aerial images to determine the vehicle's location on the images and thereby its geo-location.

Previous research in this area focuses on methods that cover large (\eg city-scale) search regions \cite{workman2015localize,liu2019lending}, but suffer from low metric accuracy \cite{zhu2022transgeo} insufficient for the navigation of autonomous vehicles. Since a prior pose estimate of the vehicle can be provided by global navigation satellite systems (GNSS) or by tracking the vehicle continuously, several recent methods employ smaller search regions to achieve higher metric accuracy\cite{shi2022beyond,fervers2022continuous}. %

Without access to three-dimensional lidar point clouds, a purely vision-based model has to bridge the gap between ground and aerial perspectives, for example by learning the transformation in a data-centric manner. We utilize a transformer model that iteratively constructs a bird's eye view (BEV) map of the local vehicle environment by aggregating information from the ground-level perspective views (PV). The BEV  refers to a nadir (\ie orthogonal) view of the local vehicle environment. The final BEV map is matched with an aerial image to predict the relative vehicle pose with three degrees of freedom (3-DoF), \ie a two-dimensional translation and a one-dimensional rotation.

Our model outperforms previous approaches for the metric CVGL task on the Ford AV \cite{agarwal2020ford} and KITTI-360 \cite{liao2021kitti} datasets and even surpasses related approaches utilizing lidar sensors in addition to camera input. It predicts a soft probability distribution over possible vehicle poses (\cf \cref{fig:intro}) rather than a single pose which specifically benefits trackers that use the model predictions to determine the vehicle's trajectory over time.

While previous works rely on the availability of training data from the target region to achieve reasonable localization accuracy, we address the strictly more challenging task of non-overlapping train and test regions. We further train and test the model on entirely different datasets that were captured with different \mbox{ground-based} vehicles. Our evaluation demonstrates the generalization capabilities of our model under cross-area and cross-vehicle conditions and highlights the potential for global-scale application without fine-tuning on a new region or a new vehicle setup.

We collect multiple datasets from the autonomous driving sector in addition to aerial images from several orthophoto providers for our evaluation. Since the vehicle's geo-locations do not always accurately match the corresponding aerial images, we compute new geo-registered ground truth poses for all datasets used in the work and filter out invalid samples via a data-pruning approach.

We publish the source code of our method online including a common interface for the different datasets. We also make the improved ground truth for all datasets publicly available. \footnote{Project page: \url{https://fferflo.github.io/projects/vismetcvgl23}}

\newenvironment{tight_enumerate}{
	\begin{enumerate}
		\setlength{\itemsep}{0pt}
		\setlength{\parskip}{0pt}
	}{\end{enumerate}}
In summary, our contributions are as follows:
\vspace{-2.2mm}\begin{tight_enumerate}
	\item We present a novel end-to-end trainable model for metric CVGL that requires only visual input and yields uncertainty-aware predictions.
	\item We collect multiple vehicle datasets and aerial images from several orthophoto providers for our evaluation. We compute improved ground truth poses using a pseudo-label approach and filter out invalid samples via data-pruning.
	\item Our method outperforms previous works by a large margin even under strictly more challenging \mbox{cross-area} and cross-vehicle settings.
\end{tight_enumerate}

\section{Related Work}
\label{sec:related}

\textbf{Cross-view Geolocalization.} CVGL refers to the task of matching camera images of a ground-based agent to geo-registered aerial images to determine the agent's geo-location. Approaches in this area typically focus on one of two problems.

\textit{Large-area CVGL} methods start from a large (\eg \mbox{city-scale}) search region and find a rough estimate of the agent's position. They typically use an image retrieval approach and therefore do not predict orientation or reach high metric accuracy (\eg less than $10\%$ of predictions reported by the \mbox{state-of-the-art} method TransGeo \cite{zhu2022transgeo} are localized with less than 10m error).

\textit{Metric CVGL} methods start from a rough pose estimate of the agent (\eg up to 50m error \cite{fervers2022continuous}) and determine the location and orientation with higher accuracy by matching the agent's sensor readings with an aerial image centered on the prior pose.

Zhu \etal \cite{zhu2021vigor} propose to differentiate the evaluation of a model into same-area and cross-area categories based on the availability of data from the test region during training. While same-area models are trained on the same aerial images used at test time and outperform cross-area models \cite{zhu2021vigor,zhu2022transgeo} they require obtaining data from the target region first. This limits their scalability and contradicts our motivation for using aerial images, \ie global availability at low cost. Furthermore, \mbox{same-area} models have not been shown to memorize large areas (\eg country-scale) from training data, even when such data is available. We therefore consider only the strictly more challenging task of cross-area CVGL.

While metric CVGL utilizing lidar and camera sensors has been shown to achieve sub-meter accurate poses \cite{fervers2022continuous}, previous purely vision-based methods have not reached comparable performance. Zhu \etal \cite{zhu2021vigor} treat the problem as a regression task on top of large-area CVGL. Xia \etal \cite{xia2021cross,xia2022visual} utilize image retrieval methods from large-area CVGL and adapt them to work with smaller search regions for metric CVGL. These methods do not explicitly consider the spatial layout of both aerial and ground input data, since the images are reduced to one-dimensional feature vectors where spatial information can only be stored implicitly in the neurons' activations.

Shi \etal \cite{shi2022beyond} propose the first method that digresses from the image retrieval paradigm of large-area CVGL by using an end-to-end differentiable Levenberg-Marquardt optimizer that iteratively estimates the relative pose between aerial and ground-level images. They rely on a flat-ground assumption by using a homography to project satellite features to the ground-level view. They further do not consider a \mbox{multi-camera} setup.

\textbf{Perspective View to Bird's Eye View}. Cameras mounted on vehicles show a two-dimensional perspective projection of the environment typically as pinhole or panoramic images. However, the preferred representation for many tasks (\eg navigation, object detection or localization), is a BEV, \ie a two-dimensional orthographic projection of the ground-level features in the top-down view. The perspective view to bird's eye view transformation (PV2BEV) represents a novel research field that has recently gained attention in the research community. \cite{ma2022vision}

PV2BEV methods can be categorized based on whether they explicitly exploit the geometry of the scene to bridge the gap between PV and BEV or learn the mapping in a data-centric manner.

Geometry-based methods typically use one of two approaches. Inverse Projective Mapping \cite{mallot1991inverse} transforms PV features to BEV via a homography based on the camera's intrinsic and extrinsic parameters \cite{ammar2019geometric,reiher2020sim2real}. These methods rely on a flat-ground assumption and are not able to properly exploit features above or below the ground-plane. \mbox{Depth-based} methods explicitly predict depth in the ground images as discrete point-clouds or probabilistic depth distributions and utilize a three-dimensional model (\eg point-cloud or voxel-map) for the projection \cite{wang2019pseudo,philion2020lift,reading2021categorical}.

Recent learning-based methods utilize transformers to project features from PV to BEV. Here, queries defined in BEV space gather information from values defined in PV space via a cross-attention mechanism. Queries are either defined sparsely (\eg for object detection) \cite{can2021structured,wang2022detr3d,liu2022petr} or as a dense spatial grid around the vehicle (\eg for semantic segmentation) \cite{peng2022bevsegformer,li2022bevformer, zhou2022cross}. Some methods further utilize deformable attention to reduce the memory consumption, such that each query attends only to a sparse set of points in the PV rather than to all PV features \cite{peng2022bevsegformer,li2022bevformer}.

\if{false}
\subsection{Placement of Our Work}

We propose a novel method for Metric CVGL that is purely vision-based and outperforms previous approaches that utilize both vision and lidar input. Our model is the first to learn the projection from PV to BEV for this task in a \mbox{data-centric} manner rather than relying on a flat-ground assumption or discarding spatial information. We use a transformer layer with deformable attention to project features from PV to BEV via a cross-attention mechanism.

Unlike previous works, we focus entirely on cross-area evaluation which allows our model to exploit the global availability of aerial images without fine-tuning on a target region. Our model further predicts soft probability distributions over possible vehicle poses which enables uncertainty estimation in the search region. We show that a simple tracker setup that utilizes the soft predictions achieves sub-meter accurate poses on KITTI-360 \cite{liao2021kitti}.
\fi

\if{false}

\paragraph{Transformer} The transformer architecture originated in the field of natural language processing \cite{vaswani2017attention} where it was developed as a model that operates on a set of tokens (\eg words in a sentence). It was later adapted to the vision domain by mapping images onto a set of tokens which are then processed by a transformer \cite{dosovitskiy2020image,touvron2021training,xie2021segformer}.

and was later adapted

- originally from nlp, works on set of tokens, then moved to Vision (ViT, Deit, Segformer, Swinformer) where input is divided into set of tokens and then processed by a transformer
- transformer block (repeated for deep networks) based on design principles:
	- main block consists of two consecutive residual blocks, token mixing (information shared between tokens) and channel mixing (information shared between channels at each token, typically MLP)
	- token mixing originally as all-to-all attention: Predict QKV for all input tokens (typically via MLP). For each query $Q_i$, aggregate information from values $V_j$ based on the similarity of the query to a value's corresponding key $K_j$. The attention map $A$ represents the weights assigned to each query-value pair and is computed as follows:
	\begin{equation}
		A = \text{softmax}(\frac{Q K^T}{\sqrt(f_{qk})})
	\end{equation}
	- alternatives to spatial mixing have been proposed: Simpler attention (Linformer etc), non-attention mixing types (Conv-mixer, MLP-mixer) => Metaformer
	- logits shortcut
	(- pre-LN vs post-LN)
	- cross-attention: Compute queries from one set of tokens and key-value pairs from another set of tokens such that for each token in the first set the model learns to gather information from the appropriate tokens in the second set.
	- we use transformer block and introduce spatial mixing operations and construction methods for queries, keys and values specifically tailored to the cross-view registration task (representing stronger inductive biases)
\fi

\begin{figure}
	\centering
	\hspace{-0.5mm}\includegraphics[width=0.85\linewidth]{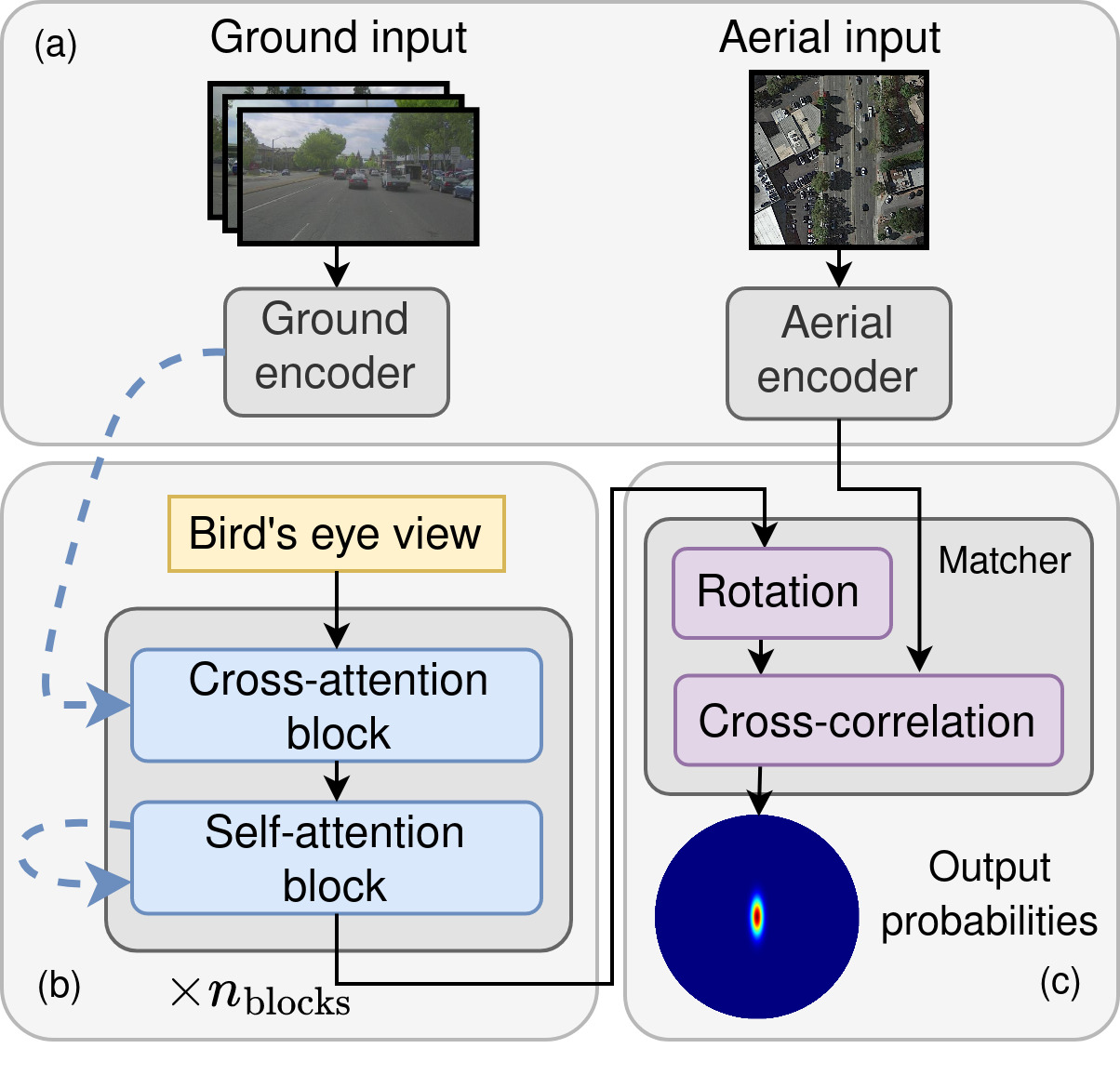}
	\vspace{-3mm}
	\caption{\textbf{Overview of the architecture.} (a) Aerial and ground input images are processed by separate encoders to predict high-level feature maps. (b) A bird's eye view (BEV) representation of the local vehicle environment is constructed. The BEV map is initialized as a grid of learnable parameters and iteratively refined by cross-attending to the ground features. (c) The final BEV and aerial features are matched via cross-correlation to predict a probability distribution over possible 3-DoF vehicle poses.}
	\vspace{-5mm}
	\label{fig:summary}
\end{figure}

\section{Method}
\label{sec:method}

Our model predicts a probability distribution for the pose of a ground-based vehicle relative to an aerial image. It first builds a BEV representation of the local vehicle environment using a cross-attention mechanism. The BEV is then matched with the aerial image to determine the vehicle's pose. \cref{fig:summary} shows a summary of the model.

\subsection{Feature extraction}

The ground and aerial input images are first processed by encoder networks that predict pixelwise features at stride $s \in \mathbb{N}$, \ie with $\frac{1}{s}$ of the original resolution. We use a shared encoder for the ground images and a separate encoder for the aerial image (\cf \cref{fig:summary}a).

We choose a lightweight architecture for the network based on common design principles as follows. For a given image, we apply a pretrained vision backbone (\ie ConvNeXt \cite{liu2022convnet}) to extract a pyramid of intermediate feature maps $L$ at strides $4, 8, 16$ and $32$. We use a global average for context pooling in the last feature map \cite{zhao2017pyramid,chen2017deeplab}. All feature maps in $L$ are resized to stride $s$ via bilinear interpolation, summed and processed by a small multilayer perceptron (MLP) to produce the final output feature map. More details are provided in the supplementary material.

The ground image of the $i$-th camera is encoded into feature map $F_{Gi}$ with $s_G \ge 1$. Similarly, the aerial image is encoded into feature map $F_A$ with stride $s_A = 1$ since its spatial resolution is particularly important for the localization accuracy. Since vision backbones typically contain a stem that initially reduces the resolution, \eg by a factor of 4, we additionally process the RGB input with two ResNet blocks \cite{he2016deep} at stride 1 and add the result to the intermediate list of feature maps $L$ to fully exploit the aerial image's spatial information. Our evaluation justifies this choice.

\subsection{Perspective View to Bird's Eye View}\label{sec:PV2BEV}

\textbf{Overview.} The BEV of the vehicle's local environment is iteratively constructed from the PVs captured by cameras at a given point in time. The BEV is centered on the vehicle and defined as a spatial grid $B \in \mathbb{R}^{d_B \times d_B \times c_B}$ with dimensions $d_B \times d_B$ (at $q_B$ meters per cell) and $c_B$ channels. It is initialized via learnable parameters and iteratively refined in $n_\text{blocks}$ steps. We regard only cells in $B$ with a distance of less than $\frac{d_B}{2} q_B$ meters to the vehicle as valid and store the corresponding mask as $M \in \{0, 1\}^{d_B \times d_B}$.

Each refinement step consists of two transformer blocks that apply cross-attention to the PV features $F_{Gi}$ as well as self-attention on $B$ (\cf \cref{fig:summary}b). In the following, the transformer block is reviewed and contrasted with the attention mechanisms used in the cross- and self-attention blocks of our model.

\begin{figure}[t]
	\centering
	\includegraphics[width=55mm]{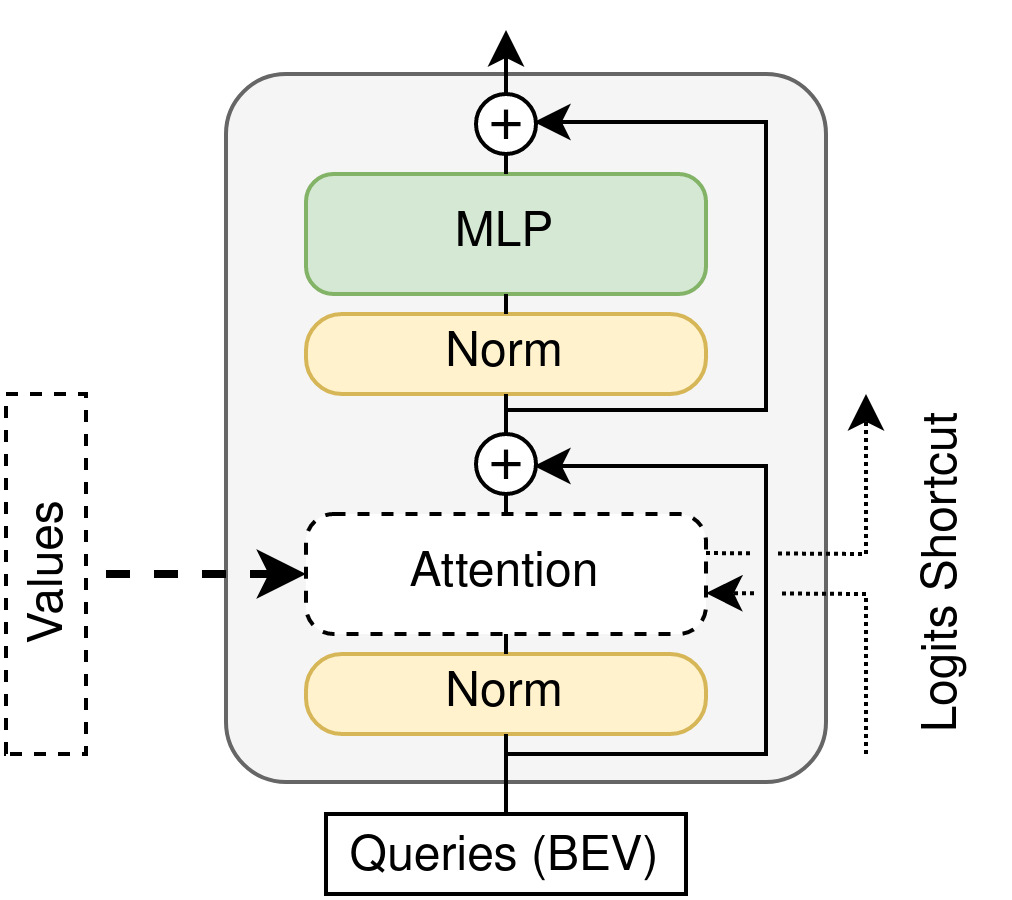}
	\vspace{-2mm}
	\caption{\textbf{Transformer block} \cite{vaswani2017attention,yu2022metaformer}. Queries representing the bird's eye view (BEV) map are used to attend to a set of values. Cross- and self-attention blocks differ based on the choice for the values (\ie PV features $F_{Gi}$ and BEV features $B$, respectively) and the type of attention mechanism that is used (\cf \cref{sec:PV2BEV}). The attention logits are optionally forwarded via a skip connection to the next attention block of the same type. Normalization is applied in the residual block (Pre-LN) \cite{xiong2020layer}.}
	\vspace{-6mm}
	\label{fig:transformerblock}
\end{figure}

The general layout of the transformer block \cite{vaswani2017attention,yu2022metaformer} is shown in \cref{fig:transformerblock}. It operates on a set of query tokens $T_q$ that adaptively aggregate information from a set of value tokens $T_{v}$. Tokens are generally packed into matrix form $T \in \mathbb{R}^{n \times c}$ with $c$ channels per token. The transformer block consists of two consecutive residual sub-blocks: The first contains an attention mechanism that distributes information from value tokens to query tokens. The second contains a MLP that processes tokens separately and is largely responsible for the representational power of the block.

The transformer blocks used in our model are based on different choices for the attention mechanism. The original design of transformers \cite{vaswani2017attention} uses query-key-value (QKV) attention for this purpose: Tokens $T_q$ are first projected onto queries $Q$, and tokens $T_v$ onto keys $K$ and values $V$ by learned linear transformations. We use the terms \textit{query} for $T_q$ and $Q$, and \textit{value} for $T_v$ and $V$ interchangeably. For each query $Q_i$, information is aggregated from values $V_j$ based on the similarity of the query $Q_i$ \wrt the value's corresponding key $K_j$. The attention map $A$ represents the weights assigned to each query-value pair and is defined as
\vspace{-2.7mm}\begin{equation} \label{eq:qkv_attention}
	A = \text{softmax}(A_\text{logit}) \hspace{5mm}\text{ with } A_\text{logit} = \frac{Q K\tran}{\sqrt{c_{qk}}}\vspace{-1mm}
\end{equation}\vspace{-0.3mm}
where $c_{qk}$ is the channel dimension of $Q$ and $K$. The values are averaged based on the weights $A$ and then linearly projected to produce the final output tokens.

For multi-head attention, the matrices $Q$, $K$ and $V$ are first split along the channel axis into $n_\text{heads}$ equally sized blocks before computation of the weighted averages. The outputs of all heads are concatenated before the final linear projection. This enables each query to incorporate $n_\text{heads}$ distinct aggregations of the input values for $n_\text{heads}$ heads.

\textbf{Cross-attention block.} The cross-attention block gathers information from the PV features $F_{Gi}$ for all cells of the BEV map $B$. Computing full cross-attention between queries $B$ and values $F_{Gi}$ such that each query attends to each PV feature leads to large memory and computational cost. Instead, for each cell in the BEV we select a small set of points in the PV and sample the features maps $F_{Gi}$ at the corresponding locations. The sampled features represent the value tokens for the corresponding BEV query token.

The PV points for a given query are determined as follows. We lift the corresponding point on the BEV into a pillar of $z$ points uniformly sampled from height $h_{\textit{min}}$ to $h_{\textit{max}}$ \cite{li2022bevformer}. The points are then transformed from vehicle coordinate system into each camera coordinate system and projected onto the camera plane using its extrinsic and intrinsic parameters. Points that do not fall into the camera frustrum are discarded. This yields up to $z$ points per query per camera. Since typical camera setups for surround view have only little overlap between cameras, most queries are assigned no more than $z$ values overall.

Since the value tokens for a given query token represent only a sparse set of points on the PV, we enable more fine-grained control over the points' locations via deformable offsets \cite{zhu2020deformable}. Given the pillar of $z$ points for a BEV query $B_{xy}$, we predict offsets $\Delta p_j \in \mathbb{R}^2$ with $j \in \{1, ..., z\}$ via $z$ learnable linear transformations on $B_{xy}$. For the $j$-th point in the pillar that is projected to location $p_{ij} \in \mathbb{R}^2$ in the $i$-th camera view, the corresponding feature $f_{ij}$ is sampled via bilinear interpolation as
\vspace{-2mm}\begin{equation}
f_{ij} = F_{Gi}(\frac{p_{ij} + \Delta p_j}{s_G})\vspace{-2mm}
\end{equation}\vspace{1mm}
where $s_G$ represents the stride of the PV feature map $F_{Gi}$. The offsets of each cross-attention block are further added onto the predicted offsets of the next block via a skip connection, such that each block learns to refine the existing offsets rather than predict entirely new offsets.

While queries are represented by a dense grid $B$, values are given as a sparse set of features $f_{ij}$ and cannot efficiently be packed into a dense spatial representation. We therefore implement values as a list of features $F_G \in \mathbb{R}^{n_v \times c_v}$ with $c_v$ channels that contains all valid features $f_{ij}$ concatenated along the first axis. The interaction between list $F_G$ and spatial grid $B$ is implemented via efficient \texttt{scatter} and \texttt{gather} operations \cite{blackford2002updated}.

To reduce the computational complexity of the transformer block, we simplify the computation of the attention map $A$ as follows. Instead of mapping query tokens $B$ and value tokens $F_G$ onto queries $Q$ and keys $K$ respectively, we predict the attention logits $A_\text{logit}$ directly from the BEV features $B$ via a learnable linear projection. Our ablation studies demonstrate that this does not lead to reduced performance. We further add a skip connection between the attention logits of subsequent blocks such that each block learns to refine the existing weights rather than predict entirely new weights \cite{he2020realformer}.

\textbf{Self-attention block.} The self-attention block refines the BEV representation via a self-attention operation on $B$. We choose a single block of the SegFormer architecture for this purpose \cite{xie2021segformer}.

In classical self-attention, $B$ is used both for query and value tokens. To avoid the large memory and computational cost of full attention, SegFormer uses spatial-reduction attention (SRA) \cite{wang2021pyramid}: While query tokens are given directly by $B$, the spatial resolution of $B$ is first reduced via a convolution with stride $s_R$ for the value tokens. This reduces the number of value tokens and thereby the computational complexity by ${s^{}_{R}}^2$. The MLP component of the self-attention block is further extended with a $3 \times 3$ depthwise convolution to mimic the use of positional encodings. Invalid features in $B$ are set to zero according to mask $M$.

\subsection{Bird's Eye View to Aerial View matching}

In the last step of the model, the aerial features $F_A$ and the final BEV features $B$ are matched to determine the relative 3-DoF pose of the vehicle (\cf \cref{fig:summary}c). We test different hypotheses \mbox{$h \in \mathcal{H} \subset \text{SE}(2)$} for the vehicle pose by comparing $F_A$ with $B$. %

In general, we choose a more fine-grained pixel resolution $q_A \le q_B$ for $F_A$ which benefits localization accuracy. Therefore, $B$ is first upsampled to match the pixel resolution $q_A$ via bilinear interpolation followed by a linear projection to $c_A$ channels.

To test a hypothesis $h$, the upsampled BEV map is transformed into aerial coordinates by the rigid transformation $h$ yielding the transformed BEV $B^{(h)}$ with the same dimensions as $F_A$. The logit of $h$ is determined as the scaled inner product of $F_A$ and $B^{(h)}$. A softmax operation is applied to the logits of all hypotheses to produce the final probability distribution as shown in \cref{eq:hypothesis_prob}. %
\vspace{-1mm}\begin{equation} \label{eq:hypothesis_prob}
P(h) = \frac{\exp k \langle F_A, B^{(h)} \rangle}{\sum\limits_{h' \in \mathcal{H}} \exp k \langle F_A, B^{(h')} \rangle} \hspace{1mm} \text{ with } k=\frac{1}{ \sqrt{\langle M, M \rangle \cdot c_A}}
\end{equation}%
Logits are scaled with $k$ to normalize the variance of the inner product as proposed by Vaswani \etal \cite{vaswani2017attention}.

We choose the set of hypotheses $\mathcal{H}_\alpha$ as a two-dimensional grid of translations around the origin with with pixel resolution $q_A$, orientation $\alpha \in \mathbb{R}$ and maximum distance $r \in \mathbb{R}$. The logits of $\mathcal{H}_\alpha$ are jointly estimated by rotating $B$ by $\alpha$ and computing the cross-correlation between $F_A$ and the rotated $B$. This is repeated for a discrete set of rotations $\alpha \in \mathcal{A}$ yielding the total set of evaluated hypotheses \mbox{$\mathcal{H} = \bigcup_{\alpha \in A} \mathcal{H}_\alpha$}. The cross-correlation is computed efficiently in the Fourier domain by utilizing the Convolution Theorem \cite{weisstein} which requires three \textit{Fast Fourier Transforms} per rotation angle per feature channel.

\subsection{Loss}

Each training sample contains the camera images, the intrinsic and extrinsic parameters, a randomly chosen apriori pose and the ground truth pose of the vehicle. We define a normal distribution $P_\text{true}$ centered on the vehicle pose with translation and angle standard deviations $\sigma_t$ and $\sigma_\alpha$ that represents the desired model output. The loss function is defined as the cross-entropy between the predicted and target probabilities:
\vspace{-2mm}\begin{equation} \label{eq:loss}
	L = - \sum_{h \in \mathcal{H}}P_\text{true}(h) \ln{P(h)}\vspace{-1mm}
\end{equation}\vspace{-2mm}

Using a soft rather than a one-hot target distribution allows training with ground truth poses that potentially lie between the discrete hypotheses. It further acts as a means for label smoothing which prevents the network from becoming \mbox{over-confident} \cite{muller2019does}.

\begin{figure}[t]
	\centering
	\begin{subfigure}[b]{\linewidth}
		\includegraphics[width=0.97\linewidth]{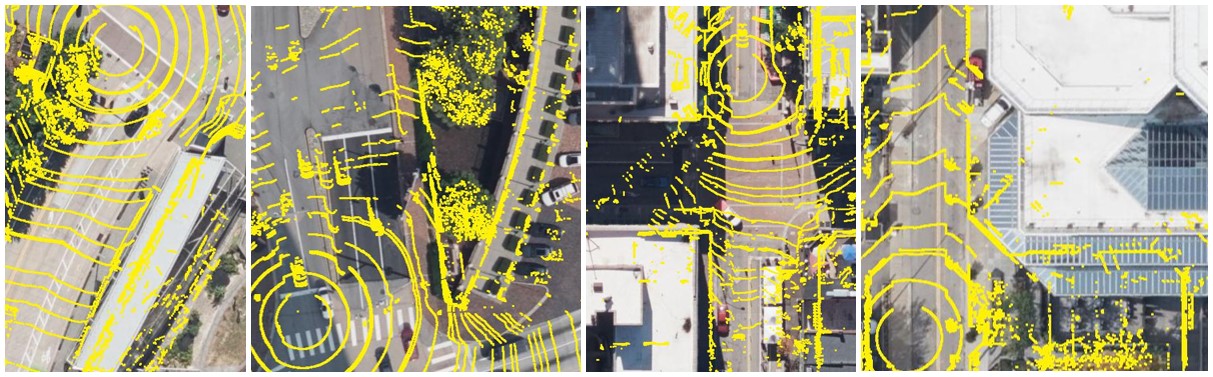}
		\caption{Original ground truth poses.}
	\end{subfigure}
	
	\begin{subfigure}[b]{\linewidth}
		\includegraphics[width=0.97\linewidth]{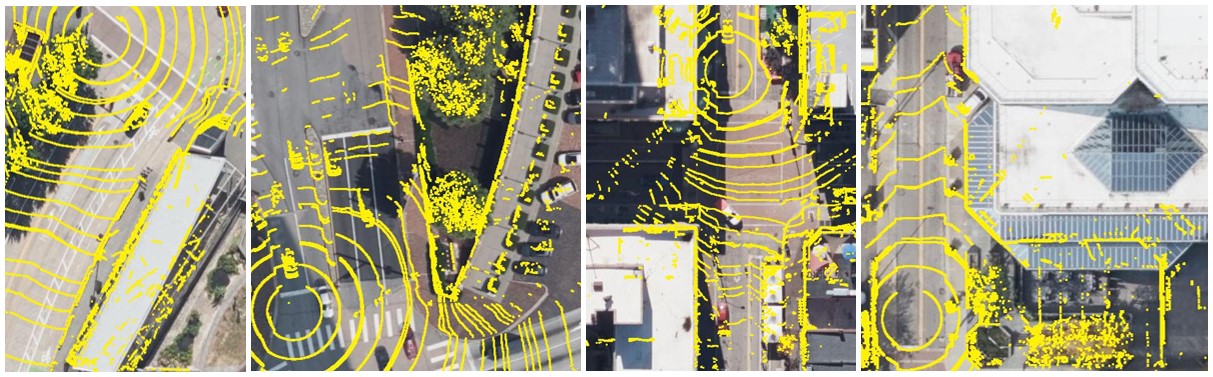}
		\caption{Our pseudo-labeled ground truth poses.}
	\end{subfigure}
	\vspace{-8mm}
	\caption{Comparison of original ground truth poses and pseudo-labeled poses. Projected lidar points are shown in yellow for visualization. Vehicle data: Argoverse V2 \cite{wilson2021argoverse}. Map data: Bing Maps © 2022 TomTom, © Vexcel Imaging \cite{bingmaps}.}
	\vspace{-5mm}
	\label{fig:pseudolabels}
\end{figure}

\section{Data}
\label{sec:data}

\textbf{Overview.} In order to build a dataset for the evaluation of CVGL methods in cross-area and cross-vehicle settings, we collect existing datasets from the autonomous driving sector (Argoverse V1 \cite{chang2019argoverse}, Argoverse V2 \cite{wilson2021argoverse}, Ford AV \cite{agarwal2020ford}, KITTI-360 \cite{liao2021kitti}, Lyft L5 \cite{houston2020one}, Nuscenes \cite{caesar2020nuscenes} and Pandaset \cite{xiao2021pandaset}) and gather aerial images for the vehicle's geo-poses from several orthophoto providers (Google Maps \cite{googlemaps}, Bing Maps \cite{bingmaps}, DCGIS \cite{dcgis}, MassGIS \cite{massgis} and Stratmap \cite{stratmap}). A detailed overview of the datasets used in this work is shown in the supplementary material. In total, the combined dataset contains 1.05 million ground \mbox{data-frames}, each consisting of the vehicle's ground truth pose and camera images and intrinsic and extrinsic parameters. This corresponds with 2.55 million pairs of ground data-frames and corresponding aerial images when counting multiple orthophoto providers.

Since subsequent frames only have a small relative offset and multiple trajectories per dataset often follow the same route, the number of paired data-frames does not reflect the data's coverage of aerial images. We measure the coverage by grouping frames into disjoint cells of size 100m $\times$ 100m which results in 5.1 thousand cells containing at least one ground frame and 13.0 thousand cells when counting multiple orthophoto providers.

We group frames into cells of size \mbox{1m $\times$ 1m} and for each training iteration randomly sample a cell from which the next frame is chosen. This prevents areas with many ground frames from being overrepresented. We resize all ground images to a minimum size of \mbox{$320 \times 240$} pixels which enables the model to run at approximately 2-3 Hz on an RTX 6000.

\begin{figure}[t]
	\centering
	\includegraphics[width=0.97\linewidth]{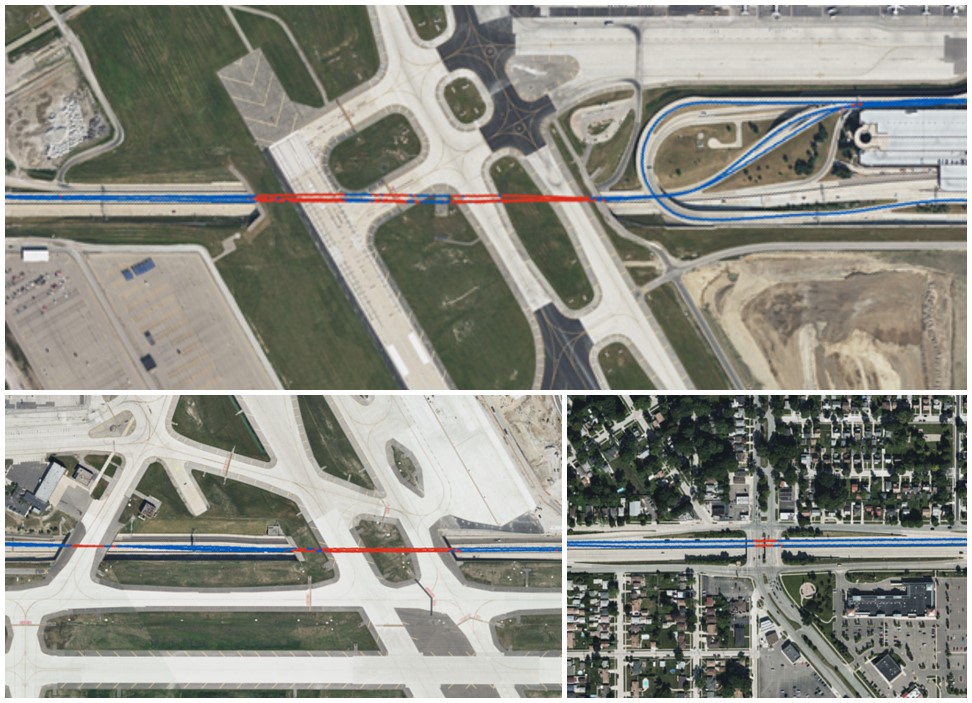}
	\vspace{-2mm}
	\caption{Examples of the data-pruning method on the Ford AV dataset \cite{agarwal2020ford}. Blue and red color represent kept and pruned frames in the trajectory. Map data: Bing Maps © 2022 TomTom, © Vexcel Imaging \cite{bingmaps}.}
	\vspace{-7.5mm}
	\label{fig:dataprune}
\end{figure}

\begin{table*}[h]
	\setlength{\tabcolsep}{4pt} %
	\newcommand{\spacing}{\rule{0pt}{3mm}}
	\def\arraystretch{0.7}
	\small
	\centering
	\caption{Recall in percent on a subset of the first two scenes of the Ford AV dataset \cite{agarwal2020ford} following the evaluation protocol introduced by Shi \etal \cite{shi2022beyond}. Results of previous works on Ford AV are provided by Shi \etal \cite{shi2022beyond}. The first three rows represent large-area CVGL methods that are adapted for metric CVGL and have no dedicated method for predicting metric offsets. Initial pose is chosen in 40m $\times$ 40m around the vehicle with up to $20^{\circ}$ of rotation noise. All methods are vision-based only.\vspace{-3mm}}
	\label{tab:ford_subset_log12}
	\begin{tabular}{c|cc|ccc|ccc|ccc|ccc|}
		& & & \multicolumn{6}{c|}{Log1} & \multicolumn{6}{c|}{Log2} \\
		& Cross- &Cross-& \multicolumn{3}{c|}{Lateral} & \multicolumn{3}{c|}{Longitudinal} & \multicolumn{3}{c|}{Lateral} & \multicolumn{3}{c|}{Longitudinal} \\
		\rule{0pt}{2.5mm} &area & vehicle & 1.0m & 3.0m & 5.0m & 1.0m & 3.0m & 5.0m & 1.0m & 3.0m & 5.0m & 1.0m & 3.0m & 5.0m \\
		\hline
		\spacing CVM-Net \cite{hu2018cvm} & \xmark & \xmark & 9.1 & 25.7 & 41.3 & 4.8 & 13.2 & 21.9 & 9.8 & 28.6 & 47.1 & 4.2 & 11.8 & 20.3 \\
		\spacing SAFA \cite{shi2019spatial} & \xmark & \xmark & 9.3 & 28.7 & 48.0 & 4.3 & 11.8 & 20.1 & 11.2 & 34.1 & 53.4 & 5.0 & 13.4 & 22.9 \\
		\spacing DSM \cite{shi2020looking} & \xmark & \xmark & 12.0 & 35.3 & 53.7 & 4.3 & 12.5 & 21.4 & 8.5 & 24.9 & 37.6 & 3.9 & 12.2 & 21.4 \\
		\hdashline
		\spacing VIGOR \cite{zhu2021vigor} & \xmark & \xmark & 20.3 & 52.5 & 70.4 & 6.2 & 16.1 & 25.8 & 20.9 & 54.9 & 75.7 & 6.0 & 16.9 & 27.0 \\
		\spacing HighlyAccurate \cite{shi2022beyond} & \xmark & \xmark & 46.1 & 70.4 & 72.9 & 5.3 & 16.4 & 26.9 & 31.2 & 66.5 & 78.8 & 4.8 & 15.3 & 25.8 \\
		
		\spacing \textbf{Ours} w/o vehicle frames & \xmark & - & 15.1 & 51.3 & 72.0 & 5.0 & 15.2 & 24.4 &11.3 & 37.8 & 62.2 & 4.7 & 15.3 & 26.0 \\
		
		\spacing \textbf{Ours} & \xmark & \xmark & \textbf{96.3} & \textbf{99.6} & \textbf{99.6} & \textbf{76.0} & \textbf{95.3} & \textbf{96.0} & \textbf{88.0} & \textbf{99.9} & \textbf{100.0} & \textbf{58.9} & \textbf{93.3} & \textbf{93.6} \\
		
		\hline
		\spacing \textbf{Ours} & \cmark & \cmark & \textbf{77.0} & \textbf{96.2} & \textbf{97.6} & \textbf{24.0} & \textbf{67.6} & \textbf{76.1} & \textbf{73.0} & \textbf{96.5} & \textbf{97.8} & \textbf{25.6} & \textbf{61.7} & \textbf{69.4} \\
	\end{tabular}
	\vspace{-6mm}
\end{table*}

\textbf{Pseudo-labels.} Similar to previous works \cite{shi2022beyond,wang2022satellite} we notice that the \mbox{geo-poses} provided by the vehicle datasets do not accurately match the corresponding aerial images and can have large relative offsets. To address this problem, we utilize a pseudo-label approach to create new \mbox{ground truth} poses for all datasets by using the lidar point-clouds contained in the data as follows.

We manually label a small subset of the data by aligning the vehicle's lidar point-clouds with the corresponding aerial images in top-down view. We train a variant of our model on this subset where the PV2BEV transformer is replaced with a simple geometric projection using the captured lidar \mbox{point-cloud} \cite{fervers2022continuous}. This model requires less data to train while still producing accurate localization results. We predict the poses for all pairs of ground frames and orthophoto providers. For each scene, the rigid transformations between subsequent frames provided by the dataset and the predicted poses and pose uncertainties are inserted into a pose graph and optimized using a least-squares approach \cite{grisetti2011g2o}. The rigid transformations are factored in with high confidence and model predictions with low confidence, such that the random (non-systematic) error in the predictions averages out over sufficiently long sequences. Our evaluation in \cref{sec:eval_kitti} and \cref{sec:ablation} justifies this approach. \cref{fig:pseudolabels} shows a qualitative comparison between original ground truth poses and pseudo-labeled poses. We publish the improved poses online to foster future research in this area.

We do not evaluate on datasets like CVUSA \cite{workman2015localize} and CVACT \cite{liu2019lending} that are typically used in large-area CVGL since a) they do not contain accurate ground truth poses and b) our pseudo-label approach cannot be utilized since the datasets do not contain lidar point-clouds or rigid transformations between frames.

\textbf{Data-pruning.}  The datasets contain samples that cannot be used for cross-view matching, \eg if the vehicle is traveling through tunnels or under bridges or if the data is \mbox{out-of-date} and does not correspond with recent aerial images. We design a simple data-pruning approach to remove these samples from the datasets as follows.

We measure the difficulty of each data-frame by processing it with the pseudo-labeling model and computing the generalized variance of the predicted probability distribution (\ie the determinant of the covariance matrix). Easy and hard samples correspond with low and high predicted variance. We sort all data-frames by their difficulty and remove the hardest $1\%$. \cref{fig:dataprune} shows examples of pruned frames on the Ford AV dataset. Our evaluation in \cref{sec:ablation} justifies this approach.

\section{Evaluation}
\label{sec:evaluation}

\begin{table*}[h]
	\setlength{\tabcolsep}{2pt} %
	\newcommand{\spacing}{\rule{0pt}{3mm}}
	\def\arraystretch{0.7}
	\small
	\centering
	\vspace{-1mm}
	\caption{Absolute Position Error in meters on KITTI-360 scenes}
	\label{tab:kitti360_ape}
	\hspace{-1mm}\begin{tabular}{c|c|c||c|c|c|c|c|c|c|c|c||c}
		Method & Camera & Lidar & 00 & 02 & 03 & 04 & 05 & 06 & 07 & 09 & 10 & Mean\\
		\hline
		\spacing Fervers \etal \cite{fervers2022continuous} & \cmark & \cmark & 0.70 & 0.94 & \textbf{0.67} & 0.95 & 0.75 & 1.16 & 0.99 & 0.75 & 2.16 & 0.94\\
		\spacing \textbf{Ours} & \cmark & \xmark & \textbf{0.62} & \textbf{0.80} & 1.01 & \textbf{0.71} & \textbf{0.62} & \textbf{0.80} & \textbf{0.60} & \textbf{0.67} & \textbf{2.12} & \textbf{0.78} \\
	\end{tabular}
	\vspace{-3mm}
\end{table*}

\subsection{Implementation details}

We use the ConvNeXt base \cite{liu2022convnet} model in the encoder for both aerial and ground images. We train for 100K iterations with the RectifiedAdam optimizer \cite{liu2019variance}, a batch size of 1 and a learning rate of $1.0 \cdot 10^{-4}$ with polynomial decay. The loss function is parametrized with $\sigma_t = 0.5 \text{m}$ and $\sigma_\alpha = 2^{\circ}$.

For the cross-attention block, we use $n_\text{heads} = 4$ heads, sample point-pillars with $z = 16$ points from $h_{\textit{min}} = -5\text{m}$ to $h_{\textit{max}} = 10\text{m}$, and encode the PV features at stride $s_G = 4$. We use a spatial reduction rate of $s_R = 4$ in the self-attention block. Overall, $n_\text{blocks} = 3$ refinement steps are applied to compute the BEV with size $d_B = 320 \text{px}$ at $q_B = 2.4 \frac{\text{m}}{\text{px}}$ and $c_B = 128$ channels. The matching is performed at resolution \mbox{$q_A = 0.3 \frac{\text{m}}{\text{px}}$} with an aerial image of size $d_A = 512 \text{px}$ and $c_A = 8$ channels.

\subsection{Per-frame evaluation}

We evaluate the per-frame performance of our model on the Ford AV dataset \cite{agarwal2020ford} and show results in \cref{tab:ford_subset_log12}. We follow the protocol introduced by Shi \etal \cite{shi2022beyond} of testing only on a subset of the first two scenes to compare our method with related approaches. The apriori pose is chosen randomly in a \mbox{40m $\times$ 40m} area around the ground truth position with up to $20^{\circ}$ of rotation noise. Since our method works with a circular search region, we choose the smallest circumscribing radius at \mbox{$\sqrt{2} \cdot 20$m}. The training is performed in three different settings as follows.

(1) \textit{Same-area and same-vehicle}: We train on the same split as Shi \etal \cite{shi2022beyond}, \ie scenes of Ford AV that are captured at a different time than the test split. Our model vastly outperforms previous approaches and successfully localizes $>90\%$ of the frames within $3m$ to the ground-truth position in the lateral and longitudinal directions.

(2) \textit{Cross-area and cross-vehicle}: We train the model on Argoverse V1, Argoverse V2, Lyft L5, Nuscenes and Pandaset, but remove data from Detroit where Ford AV was recorded. During training, the model therefore does not have access to data from either the test region (\ie Detroit) or the test vehicle and corresponding camera setup (\ie Ford AV vehicle). Our model still outperforms previous approaches trained in the same-area and same-vehicle setting by a large margin.

\cref{fig:intro} shows several examples of probability distributions predicted by our model on the Ford AV dataset under \mbox{cross-area} and cross-vehicle conditions. With a search radius of 50m, we achieve a median position error over all six trajectories of 0.87m both when the orientation is known (up to $30^{\circ}$ noise) and when no orientation information is given. We provide more details per trajectory in the supplementary material.

\begin{figure}[t]
	\centering
	\begin{subfigure}[b]{\linewidth}
		\includegraphics[width=1\linewidth]{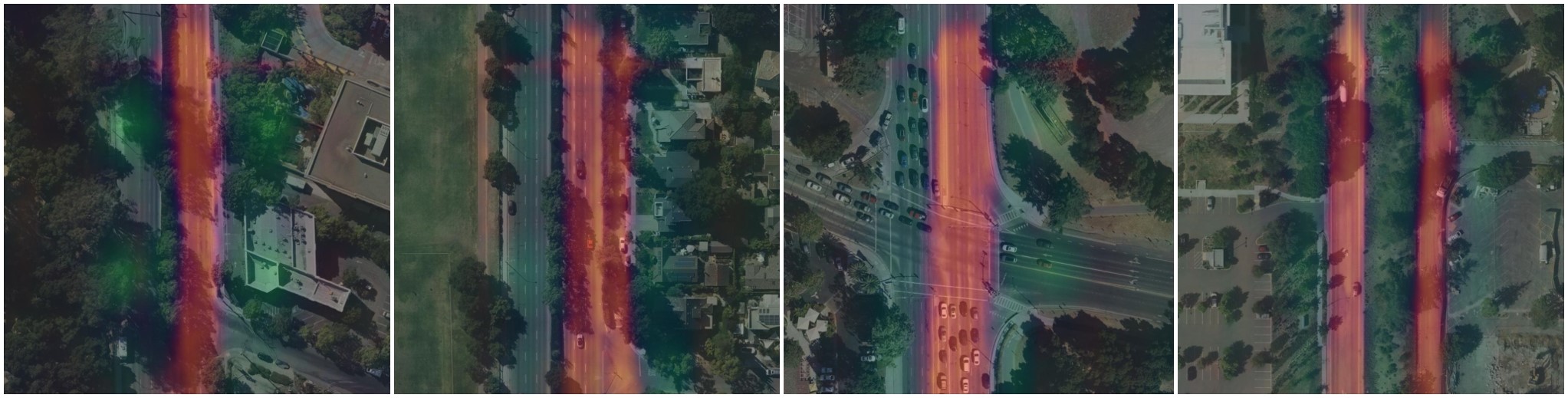}
		\caption{Model predictions with \textit{known} orientation. Driving direction points upwards. In the first three examples the model exploits its learned knowledge of \mbox{right-hand} side traffic in the training data.}
	\end{subfigure}
	
	\begin{subfigure}[b]{\linewidth}
		\includegraphics[width=1\linewidth]{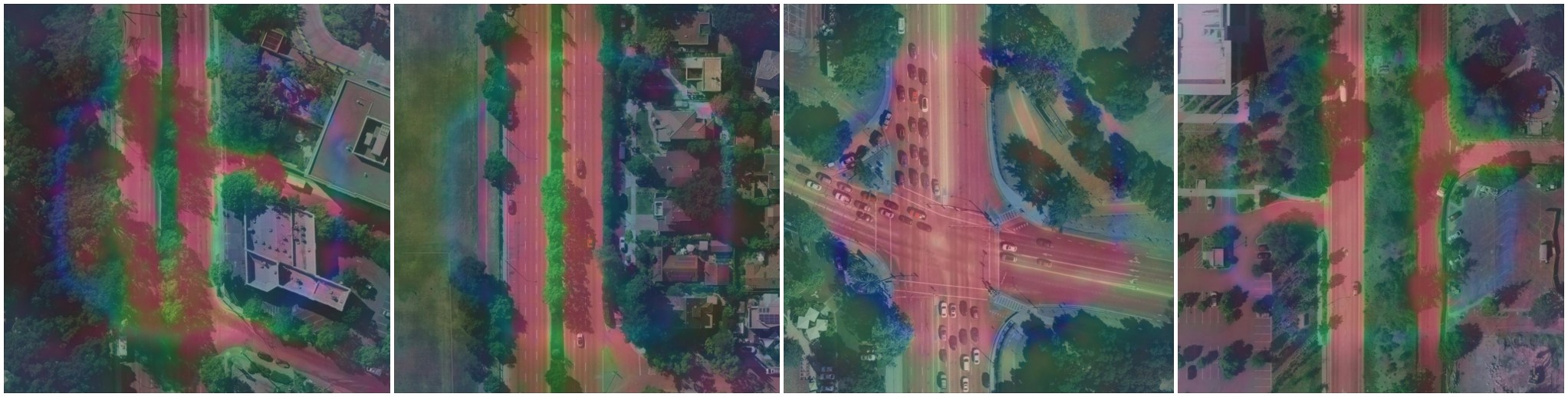}
		\caption{Model predictions with \textit{unknown} orientation.}
	\end{subfigure}
	\vspace{-5mm}
	\caption{Aerial features predicted by a model that was trained in a cross-area setting without ground cameras and therefore learns only a prior distribution of vehicle poses \wrt the aerial image. Features are reduced to three channels via principal component analysis and mapped onto RGB for visualization. Vehicle data: Lyft L5 dataset \cite{houston2020one}. Map data: Bing Maps © 2022 TomTom, © Vexcel Imaging \cite{bingmaps}.}
	\vspace{-4.5mm}
	\label{fig:driveseg}
\end{figure}

(3) \textit{Same-area and no-vehicle}: We additionally train our model without using information from the vehicle cameras, \ie by setting all RGB values to zero. The model therefore only learns a prior distribution of vehicle poses \wrt the aerial image since the BEV map is constant over different inputs. The model shows a performance similar to the previous state-of-the-art \mbox{HighlyAccurate} \cite{shi2022beyond} for longitudinal recall indicating that their model might rely mainly on prior poses \wrt the aerial image rather than on cross-view matching (\cf \cref{tab:ford_subset_log12}). \cref{fig:driveseg} shows a visualization of features learned by the corresponding model.

\subsection{Tracking evaluation}
\label{sec:eval_kitti}

We choose a tracking method using a Kalman filter to determine the trajectory of the vehicle over time \cite{fervers2022continuous}. Here, an inertial measurement unit is used to produce accurate short-term trajectories while the predictions of our model keep the position in alignment with aerial images in the long term. The tracker particularly benefits from the model's uncertainty estimates which are fed into the Kalman filter and propagated over time.

We test our model with this tracking method on the \mbox{KITTI-360} \cite{liao2021kitti} dataset and train in a cross-area and \mbox{cross-vehicle} setup on all other datasets. To ensure fair comparison with related works, rather than measuring the error \wrt our pseudo-labels we choose an evaluation approach which is typically used for odometry methods and is able to compensate for relative offsets between ground truth \mbox{geo-locations} and aerial images \cite{Zhang18iros}. We align the predicted trajectories with the original ground truth of KITTI-360 via a 3-DoF rigid transformation and measure the relative deviation of the transformed trajectories to the ground truth. As shown in \cref{tab:kitti360_ape}, this method has a mean position error of 0.78m over all scenes which surpasses even a recent lidar-visual based work.

The error \wrt our pseudo-labels (without prior alignment) is 0.85m which supports the quality of our \mbox{pseudo-labeled} ground truth. We further provide two videos for the tracking results of scenes from KITTI-360 and \mbox{Ford AV} in the supplementary material for a qualitative evaluation.

\subsection{Ablation studies}
\label{sec:ablation}

\begin{table}[t]
	\setlength{\tabcolsep}{4pt} %
	\newcommand{\spacing}{\rule{0pt}{3mm}}
	\def\arraystretch{0.85}
	\small
	\centering
	\centering
	\caption{Ablation studies tested on all scenes from Palo Alto and San Francisco. The initial pose is chosen randomly up to 30m from the vehicle with up to $10^{\circ}$ of rotation noise. ME: Mean error in meters. RMSE: Root mean squared error in meters.\vspace{-2mm}}
	\label{tab:ablation}
	\begin{tabular}{c|c| c}
		Method modification & ME & RMSE \\
		\hline
		\spacing -- & 1.19 & 3.44 \\
		No deformable offsets & 1.22 & 3.65 \\
		No ResNet blocks at stride 1 & 1.22 & 3.51 \\
		$n_\text{heads} = 1$ & 1.23 & 3.59 \\
		No deformable offset skip connection & 1.23 & 3.69 \\
		No data pruning & 1.23 & 3.62 \\
		No MLPs in transformer blocks & 1.23 & 3.52 \\
		$n_\text{blocks} = 1$ & 1.24 & 3.76 \\
		No attention skip connection & 1.24 & 3.77 \\
		AV $\rightarrow$ QKV attention & 1.25 & 3.63 \\
		No Self Attention Block & 1.38 & 4.21 \\
		No pseudo-labels & 2.37 & 5.15 \\
		No BEV upsampling & 2.63 & 5.87 \\
		No encoder pretraining & 4.15 & 9.21 \\
		No vehicle images & 11.95 & 15.20 \\
	\end{tabular}
	\vspace{-5mm}
\end{table}

For the ablation studies, we choose a smaller model with the nano variant of ConvNeXt \cite{rw2019timm} as encoder, a BEV map with size $d_B = 192 \text{px}$ at $q_B = 2.0 \frac{\text{m}}{\text{px}}$ and $c_B = 32$ channels, and aerial features with size $d_B = 256 \text{px}$ at $q_A = 0.5 \frac{\text{m}}{\text{px}}$. We use all scenes from Palo Alto and San Francisco as test split and train on the rest excluding KITTI-360 which does not contain full surround view with cameras.

We remove the individual components of our method listed in \cref{tab:ablation} and report the corresponding test scores to evaluate their effect on the localization accuracy. All components improve the performance of the model which supports their motivation in \cref{sec:method} and \cref{sec:data}.

\section{Conclusion}
\label{sec:conclusion}

We present a novel method for vision-based cross-view geolocalization that allows localizing a vehicle on an aerial image with high metric accuracy. To evaluate the method in cross-area and cross-vehicle settings, we combine multiple vehicle datasets with aerial images from several orthophoto providers to train and test our method. We implement a pseudo-label approach to improve the inaccurate ground truth poses of these datasets, and make the improved ground truth publicly available. Our method outperforms previous approaches by a large margin even under more challenging cross-area and cross-vehicle conditions. We further show that a standard tracking framework is capable of exploiting the soft probability distributions predicted by our model to determine the vehicle's trajectory over time with sub-meter accurate poses.

{\small
\bibliographystyle{ieee_fullname}
\bibliography{11_references}
}

\end{document}